\documentclass[10pt,twocolumn,letterpaper]{article}

\usepackage{iccv}              
\usepackage{color}
\usepackage{multirow}
\usepackage[accsupp]{axessibility}
\definecolor{camera}{rgb}{0.0, 0.0, 0.0}

\definecolor{iccvblue}{rgb}{0.21,0.49,0.74}
\usepackage[pagebackref,breaklinks,colorlinks,allcolors=iccvblue]{hyperref}

\def\paperID{8389} 
\def\confName{ICCV}
\def\confYear{2025}

\title{Dark-ISP: Enhancing RAW Image Processing for Low-Light Object Detection}

\author{Jiasheng Guo$^{1,}$\thanks{These authors contributed equally to this work.} \qquad Xin Gao$^{1,}$\footnotemark[1] \qquad  Yuxiang Yan$^{1}$ \qquad  Guanghao Li$^{1}$ \qquad  Jian Pu$^{1,}$\thanks{Corresponding author.}
\\
$^{1}$ Institute of Science and Technology for Brain-inspired Intelligence, Fudan University\\
{\tt\small \{guojs22, gaoxin23, yxyan22, ghli22\}@m.fudan.edu.cn},\quad
{\tt\small \{jianpu\}@fudan.edu.cn}
}

\begin{document}
\maketitle
\begin{abstract}

Low-light Object detection is crucial for many real-world applications but remains challenging due to degraded image quality. While recent studies have shown that RAW images offer superior potential over RGB images, existing approaches either use RAW-RGB images\footnote{In this paper, we use raw Bayer sensor data directly, while some methods quantize high-bit-depth `Bayer RAW' into `RAW-RGB' images with 8 bit-depth\cite{chen2023instance, cui2024raw}. We distinguish between these two terms.} with information loss or employ complex frameworks. 
To address these, we propose a lightweight and self-adaptive Image Signal Processing (ISP) plugin, \textbf{Dark-ISP}, which directly processes Bayer RAW images in dark environments, enabling seamless end-to-end training for object detection. Our key innovations are: 
(1) We deconstruct conventional ISP pipelines into sequential linear (sensor calibration) and nonlinear (tone mapping) sub-modules, recasting them as differentiable components optimized through task-driven losses. Each module is equpped with content-aware adaptability and physics-informed priors, enabling automatic RAW-to-RGB conversion aligned with detection objectives.
(2) By exploiting the ISP pipeline’s intrinsic cascade structure, we devise a Self-Boost mechanism that facilitates cooperation between sub-modules. 
Through extensive experiments on three RAW image datasets, we demonstrate that our method outperforms state-of-the-art RGB- and RAW-based detection approaches, achieving superior results with minimal parameters in challenging low-light environments.
\end{abstract}    
\section{Introduction}
\label{sec:intro}


Object detection in low-light conditions is vital for applications such as autonomous driving and surveillance \cite{girshick2015fast, lin2017focal, redmon2016you, carion2020end}. However, darkness induces severe image degradation characterized by noise amplification and diminished contrast, thereby posing substantial challenges to detection algorithms. Traditional methods using  compressed dynamic
range RGB images are limited by the low-bit-depth information and noise introduced during image signal processing (ISP), significantly reducing their effectiveness in low-light environments \cite{seo2020embeddedpigdet, xu2021exploring, lin2023dark, xue2022best}. 

Unlike RGB images, RAW images directly capture sensor orignal data before ISP processing, preserving physically meaningful information such as scene radiance and noise characteristics \cite{wei2020physics, Wei_Fu_Zheng_Yang_2021}. This results in RAW images containing richer details and color information compared to RGB images, offering greater flexibility in post-processing. Furthermore, RAW format files also store camera metadata during the photography process, which can serve as auxiliary information to provide additional guidance for deep network training. 


\begin{figure}[t]
\centering
\includegraphics[width=1\columnwidth]{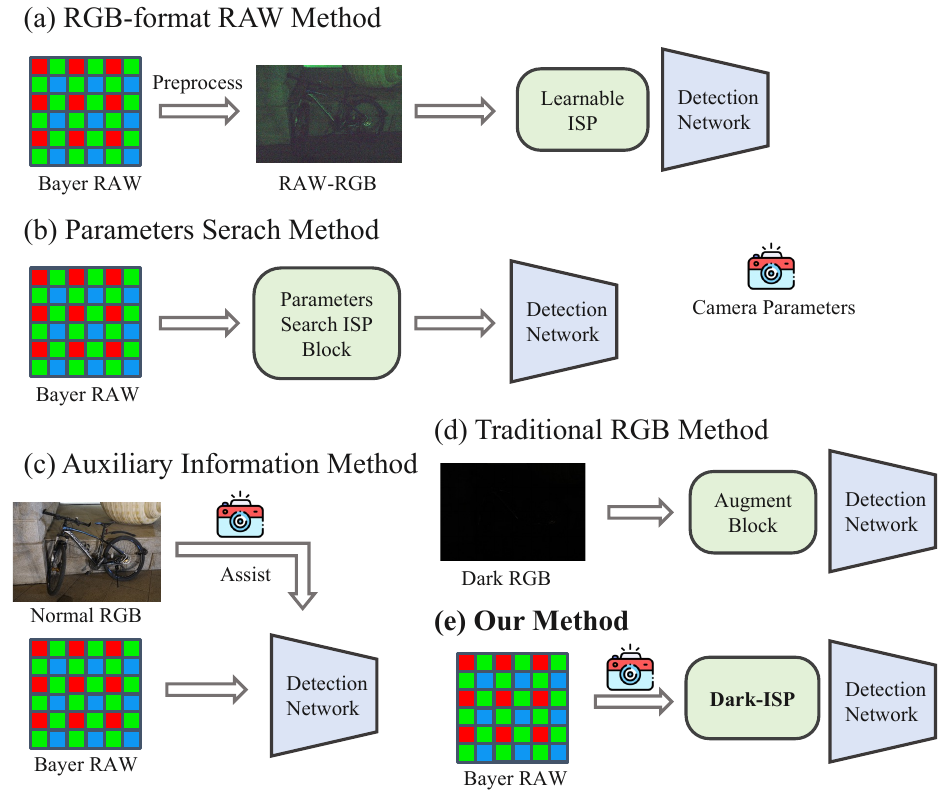} 
\caption{Thumbnails of different low-light object detection pipelines. (a) Methods utilizing RAW-RGB data. (b) ISP parameter search methods. (c) Methods incorporating auxiliary information for training. (d) Dark RGB image enhancement pipeline. (e) Our proposed Dark-ISP.}
\label{fig:comparsion}
\end{figure}

Effectively and efficiently utilizing the rich information in RAW images for object detection remains a significant challenge. Existing methods to bridging the RAW image and downstream networks pretrained on RGB domain can be broadly classified into three categories.
The first approach (\cref{fig:comparsion}(a)) bining and quantize four-channel Bayer RAW images into three-channel RAW-RGB to ensure compatibility with pretrained backbone, as seen in LIS \cite{chen2023instance}, Crafting \cite{hong2021crafting}, and RAW-Adapter \cite{cui2024raw}. However, this conversion reduces bit-depth and leads to substantial information loss, limiting the benefits of RAW data.
The second approach (\cref{fig:comparsion}(b)) incorporates parameters searching ISP, such as NOD \cite{morawski2022genisp}, AdaptiveISP \cite{wang2025adaptiveisp}, Qin \textit{et al.}\cite{qin2022attention}, and DynamicISP \cite{yoshimura2023dynamicisp} which directly process Bayer RAW images using trainable ISP modules to bridge the RAW and RGB domains. While effective, these methods often introduce computationally expensive processes, such as complex parameter searching algorithms and multi-stage training strategy, making them impractical for real-world deployment.
The third category (\cref{fig:comparsion}(c)) relies on auxiliary information augmentation, where methods like ISP-Teacher \cite{zhang2024isp} and Multitask AET \cite{cui2021multitask} integrate normal-light RGB or camera metadata references during training, but the requirement for such additional data may not always be necessary.

To address these limitations, we propose \textbf{Dark-ISP}, a lightweight and self-adaptive ISP plugin that fully exploits the advantages of RAW images for low-light object detection. By analyzing the sequential steps of camera ISP, we design a decoupled framework consisting of linear and nonlinear components.
The \textbf{linear component} captures fundamental camera operations through a sequence of matrix operations, enhanced by Local-Global Attention to dynamically balance sensor-derived processes with scene patterns.
For the \textbf{nonlinear component}, inspired by the tone mapping function’s ability to stretch dark regions and compress bright areas, we define a set of non-convex polynomial bases with strong physical interpretability. Instead of learning arbitrary transformations, our method learns the coefficients to combine these bases into an effective nonlinear operation. 
To encourage synergy between these two sub-modules, we also introduce a \textbf{Self-Boost Regularization} that uses the output of the nonlinear component to guide the learning direction of the linear mapping. Extensive validation across both real-world and synthetic RAW datasets demonstrates that our approach outperforms previous RGB-RAW methods, exhibiting superior generalization ability. 

The main contributions of this work are as follows: 
\begin{itemize} 
\item We propose a lightweight ISP plugin that separates linear and nonlinear processing, integrating content-aware adaptability and physics-informed priors to fully exploit Bayer RAW image for low-light object detection.
\item We propose a Self-Boost mechanism that enhances synergy between sub-modules in the ISP, improving robustness across varying lighting conditions.
\item We demonstrate superior performance over existing methods with fewer parameters on three low-light RAW image object detection datasets. 
\end{itemize}

\section{Related Work}
\label{sec:related work}

\subsection{Low-light Object Detection}

Low-light object detection remains a challenging task due to weak  illumination  and noise degradation. Existing methods can be broadly categorized into image-level and feature-level approaches.
\textit{Image-level methods} process images prior to detection through illumination enhancement \cite{xu2021exploring, guo2021dynamic} and noise suppression \cite{hashmi2023featenhancer, peng2024novel}. 
For example, FeatEnHancer \cite{hashmi2023featenhancer} hierarchically combines multiscale features to enhance the image before downstream network.
\textit{Feature-level methods} focus on designing specialized modules and learning strategies to extract clean, informative features \cite{lin2023dark, du2024boosting}.
\textit{Hybrid approaches} \cite{peng2024novel, cui2024trash, xue2022best} integrate the above two categories into a unified framework, enabling synergistic interaction. 
For instance, contrastive learning \cite{xue2022best} jointly optimizes both components to enhance image quality and detection performance.
While existing works have achieved promising results, they predominantly operate on processed RGB images that inherently suffer from information loss and compound noise introduced in the ISP. Recent attempts \cite{chen2023instance, hong2021crafting} use  quantized RAW data, which reduces bit-depth and loses sensor-specific information, making them less robust. We argue that \textit{unquantized Bayer RAW images} retain superior photonic information and dynamic range, motivating our framework to directly leverage RAW sensor data for end-to-end low-light object detection and bypass the information bottleneck of RGB-based approaches.



\begin{figure*}[t]
\centering
\includegraphics[width=2\columnwidth]{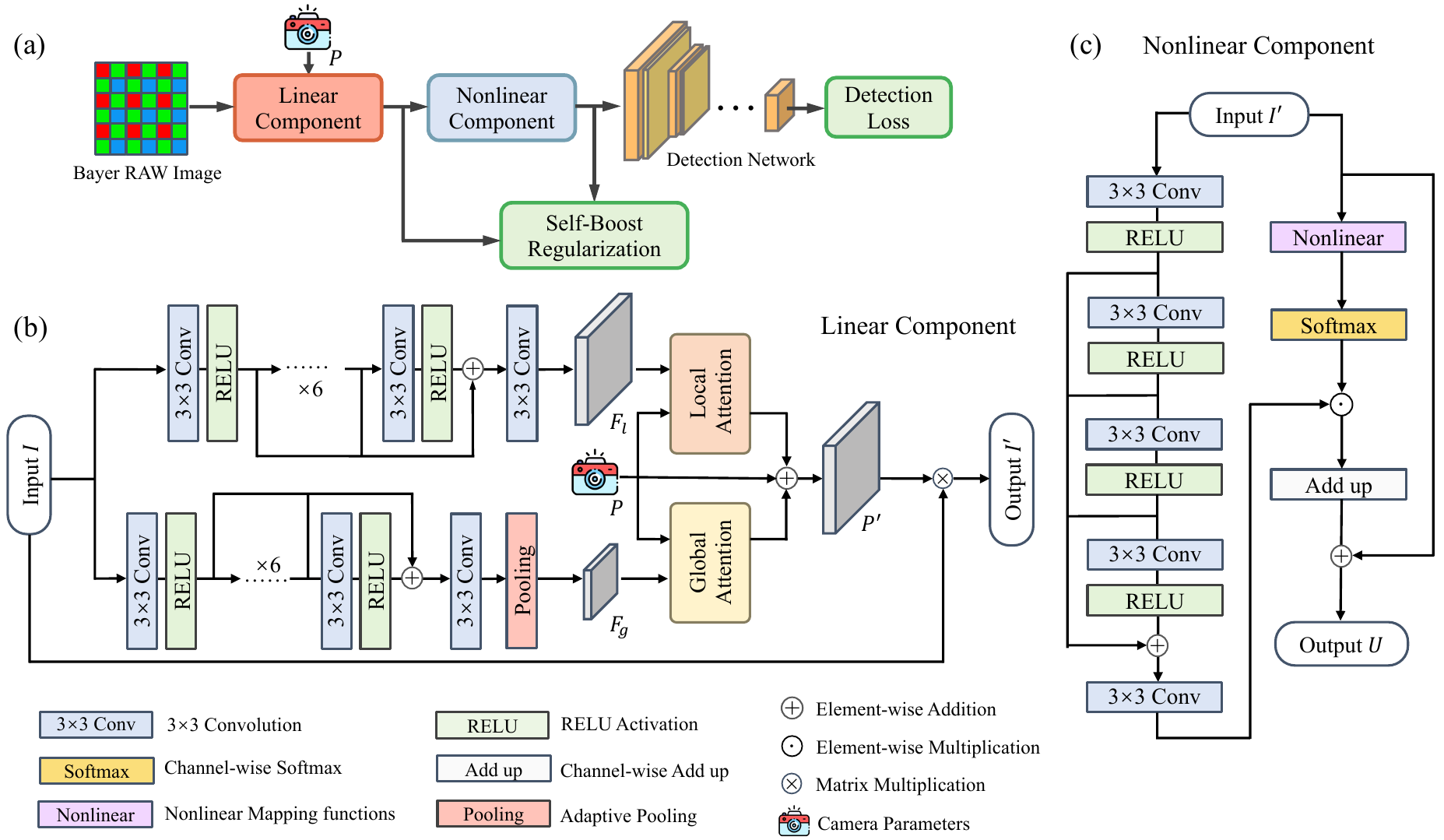} 
\caption{(a) Overview of Dark-ISP. The linear and nonlinear components process Bayer-format inputs $I$ sequentially, with the Self-Boost Regularization $\mathcal{L}_{sb}$ guiding the learning of the linear component. (b) \textbf{Linear
Component}: $P$ is the camera parameter matrix. $F_l$ and $F_g$ represent the local and global features, which produce context-aware matrices via the Attention mechanism, combined to form an adaptive linear transformation $P'$. (c) \textbf{Nonlinear
Component}: The image $I'$ from the Linear Component is passed through a network to generate pixel-wise coefficient maps, combined with predefined bases for the nonlinear transformation.}
\label{fig:framework}
\end{figure*}

\subsection{Image Signal Processing}

Image Signal Processing (ISP) transforms RAW sensor data into high-quality images. Traditional ISP methods rely on manually designed components executed sequentially \cite{lee2011local, zhu2011restoration, eilertsen2015real, delbracio2021mobile}. Recently, deep learning has been incorporated into ISP, with two main strategies:
\textit{Image-to-image frameworks} focus on learning the mapping from RAW to RGB images. Differentiable modules have been introduced into the ISP pipeline, enabling end-to-end optimization for high-quality image generation \cite{li2024dualdn, kim2023paramisp, guo2020zero, ljungbergh2023raw}. Symmetric bidirectional structures \cite{zamir2020cycleisp, xing2021invertible, brooks2019unprocessing} also have been proposed to learn reversible mappings between RAW and sRGB images, facilitating reversible conversion and preserving more information from the original RAW data. 
To improve color manipulation, lookup tables (LUTs) \cite{conde2024nilut, yang2022seplut, zeng2020learning, wang2021real} have been integrated with deep networks to efficiently retrieve pixel color values through lookup and interpolation operations.
\textit{Task-driven ISP optimization} directly adjusts ISP parameters based on task-specific objectives. GenISP \cite{morawski2022genisp} and Guo \textit{et al.} \cite{guo2024learning} adopt multi-stage training to align ISP outputs with downstream network requirements. Several methods refine ISP components using reinforcement learning for adaptive search \cite{wang2024adaptiveisp} or gradient-based optimization guided by task losses \cite{qin2022attention, yoshimura2023dynamicisp, mosleh2020hardware}. ISP-teacher \cite{zhang2024isp} improves robustness by leveraging well-lit RGB images as auxiliary inputs. RAW-Adapter \cite{cui2024raw} introduces learnable queries to refine ISP parameters and enable interaction with downstream networks. Compared to these studies, our design streamlines both training and deployment through its straightforward architecture.



\subsection{Low-Light RAW Image Enhancement}

The potential of RAW images for low-light enhancement was first highlighted by SID \cite{chen2018learning}, sparking subsequent research to bypass the RGB domain and directly process RAW images. Some methods \cite{wei2020physics, jin2023lighting, cao2023physics, feng2022learnability, zhang2021rethinking, zhang2023towards} develop noise models for dark RAW images, replacing traditional Gaussian models with more accurate CMOS sensor-based approaches for noise removal. Other works \cite{chen2019seeing, jin2023dnf, lin2023unsupervised, diamond2021dirty, dong2022abandoning} utilize heuristic frameworks and advanced network designs \cite{chen2023tsdn, chen2023masked, xu2022rawformer} to directly map noisy to clean RAW images. Recently, techniques like Diffusion \cite{dagli2023diffuseraw, wang2023exposurediffusion} and Mamba \cite{chen2024retinex} have also been explored for RAW image enhancement. However, these approaches focus primarily on image-level enhancement without integrating downstream tasks. In contrast, our method connects RAW images directly to downstream object detection tasks, enabling end-to-end training through a mediating ISP structure that bridges the RAW and RGB domains.


\section{Method}

We present the Dark-ISP framework in \cref{fig:framework}. Following the conventional camera ISP pipeline, the RAW images first pass through the linear mapping module (\cref{method:sec1}), which enhances standard camera operations with a context-aware attention mechanism. The nonlinear component (\cref{method:sec2}) then applies a set of concave function bases to model effective nonlinear transformations. To promote synergy between these components, we introduce a Self-Boost Regularization loss (\cref{method:sec3}) that aligns their optimization.



\subsection{Dynamic Linear Mapping}\label{method:sec1}


In traditional ISP pipelines, operations like \textit{White Balance}, \textit{Binning}, and \textit{Color Space Transform} are performed sequentially as linear processes \cite{delbracio2021mobile}. These steps adjust color balance, convert Bayer images \( I \in \mathbb{R}^{4 \times H \times W} \) to RGB, and map RAW sensor data to a perceptually uniform color space using matrix transformations.

First, the \textit{White Balance} operation compensates for light source variations by applying a diagonal gain matrix \( W \in \mathbb{R}^{4 \times 4} \) to the RAW data:
\begin{equation}\begin{scriptsize}
  \begin{pmatrix}
r' \\
g'_r \\
b' \\
g'_b
\end{pmatrix} 
= 
\begin{pmatrix}
 w_{1} & 0 & 0 & 0 \\
0 &  w_{2} & 0 & 0 \\
0 & 0 &  w_{3} & 0 \\
0 & 0 & 0 &  w_{4} \\
\end{pmatrix} 
\cdot
\begin{pmatrix}
r \\
g_r \\
b \\
g_b \\
\end{pmatrix}.\end{scriptsize}
  \label{eq:1}
\end{equation}
Next, the \textit{Binning} operation merges the four Bayer channels into three RGB channels, averaging the two green channels using the sparse matrix $B \in \mathbb{R}^{3 \times 4}$:
\begin{equation}\begin{scriptsize} 
  \begin{pmatrix}
R \\
G \\
B \\
\end{pmatrix} 
= 
\begin{pmatrix}
1 & 0 & 0 & 0\\
0 & \frac{1}{2} & 0 & \frac{1}{2} \\
0 & 0 & 1 & 0 \\
\end{pmatrix} 
\cdot
  \begin{pmatrix}
r' \\
g'_r \\
b' \\
g'_b
\end{pmatrix} .\end{scriptsize}
  \label{eq:2}
\end{equation}
Following this, the \textit{Color Space Transform} step uses a color correction matrix \( C \in \mathbb{R}^{3 \times 3} \) to map the image to a perceptual color space: \cite{delbracio2021mobile}:
\begin{equation}
\begin{scriptsize} 
  \begin{pmatrix}
R' \\
G' \\
B' \\
\end{pmatrix} 
= 
\begin{pmatrix}
c_{11} & c_{12} & c_{13} \\
c_{21} & c_{22} & c_{23} \\
c_{31} & c_{32} & c_{33}  \\
\end{pmatrix} 
\cdot
  \begin{pmatrix} 
R \\
G \\
B \\
\end{pmatrix} .\end{scriptsize}
  \label{eq:3}
\end{equation}
The entire process can be compactly summarized as a single matrix $P \in \mathbb{R}^{3 \times 4}$ that transforms the RAW image $I$ into the corrected RGB image $I' \in \mathbb{R}^{3 \times H \times W}$:
\begin{equation}
  I' = C\cdot B \cdot W \cdot I
  = P \cdot I .
  \label{eq:4}
\end{equation}
The white balance coefficients and the CCM are typically determined through automatic white balance algorithms in the camera, as well as manual regression estimates against standard colors \cite{delbracio2021mobile}. Since these parameters are specific to each camera, they remain largely static within the image processing pipeline. To improve the communication between the image, downstream tasks, and these parameters, we treat them as part of the learning process.

To enhance adaptability across sensors and low-light conditions, we introduce a dynamic mapping. Inspired by hierarchical feature extractors \cite{Cai_Vasconcelos_2018, Zhu_Su_Lu_Li_Wang_Dai_2020, sun2021sparse, zhang2022featurized}, the RAW image \( I \) is first processed through a dual-stream architecture to extract pixel-level local features \( F_l \in \mathbb{R}^{C \times H \times W} \) and image-level global features \( F_g \in \mathbb{R}^{C \times \frac{H}{16} \times \frac{W}{16}} \), which are then used in Local and Global Attention mechanisms \cite{vaswani2017attention} to generate pixel-level and image-level operations:
\begin{equation}
  \begin{aligned}
    P_l &= \text{LocalAttn}(Q=F_l, K=P, V=P) \in \mathbb{R}^{(3 \times 4)\times H \times W} \\
    P_g &= \text{GlobalAttn}(Q=P, K=F_g, V=F_g) \in \mathbb{R}^{3 \times 4}.
  \end{aligned}
  \label{eq:5}
\end{equation}
The local and global operations are then combined using an arithmetic sum (broadcasting) with a skip connection to produce the adaptive linear transformation $P' = (P_l + P_g + P) \in \mathbb{R}^{3 \times 4 \times H \times W}$. These parameters are applied to the RAW image to produce the corrected RGB output:
\begin{equation} 
I' = (P_l + P_g + P) \cdot I = P' \cdot I. \label{eq:7} 
\end{equation} 
This process is driven by both photographic principles and task-specific feedback, enabling a flexible, content-aware transformation that enhances performance under low-light conditions, particularly for object detection.

\begin{figure}[t]
\centering
\includegraphics[width=0.75\columnwidth]{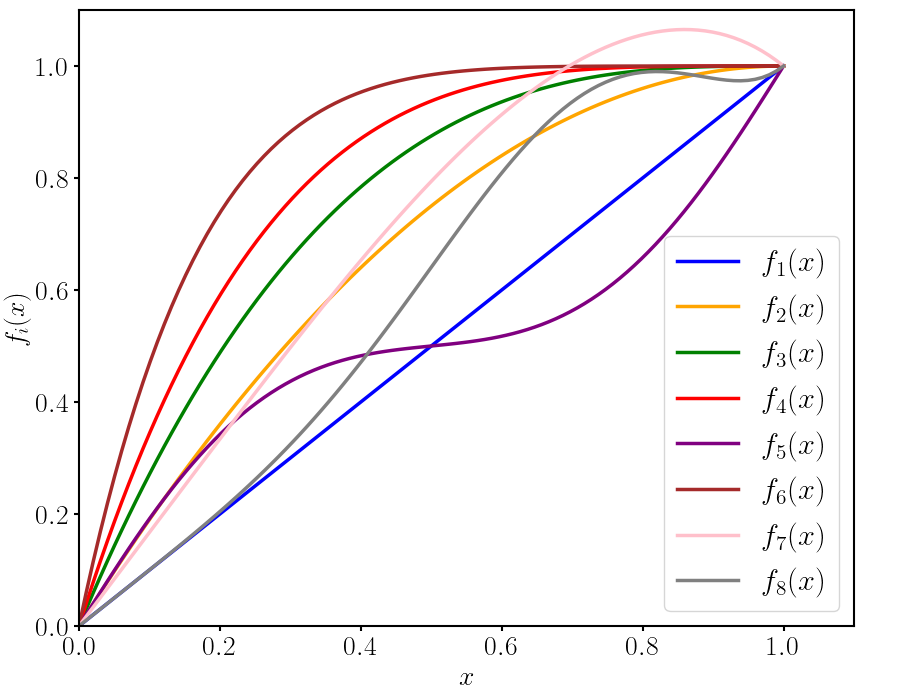} 
\caption{Non-convex polynomial bases from first-order to eighth-order. $f_k$ represents the polynomial of order $k$.}
\label{fig:nolinear}
\end{figure}



%


\subsection{Nonlinear Stretch with Polynomial Bases}\label{method:sec2}

The nonlinear component of the Image Signal Processor (ISP) is crucial for adjusting the RGB image’s color distribution. It enhances visual aesthetics and supports pixel-wise operations such as denoising and sharpening. Rather than relying on a black-box neural network transformation, we explicitly design \textbf{physically interpretable function bases} to ensure the desired properties and adaptively combine them for image-specific enhancement.

For each image \( I' \) produced by the linear module, a network predicts pixel-wise coefficients \( \{C_k\}_{k=0}^n \). These coefficients are then combined with a set of polynomial bases \( \{f_k(x)\}_{k=0}^n \) to form the nonlinear transformation:
\begin{equation}
    \mathcal{F}(x_{ij})=\sum_{k=0}^n C_k(i,j) f_k(x_{ij}).
\end{equation} 
Here, \( (i,j) \) represents the pixel position. This transformation can approximate a complex, locally smooth function \( \mathcal{H}(\cdot) \) with high-order infinitesimal errors \( o(x^{n}) \):
\begin{equation}
\mathcal{H}(x) = \mathcal{F}(x) + o(x^{n}) = \sum_{k=0}^n C_k(i,j) f_k(x) + o(x^{n}).
\end{equation}
This is analogous to an \( n \)-order Taylor expansion, where terms are reordered and coefficients combined.

To design the bases that govern the nonlinear transformation, we focus on addressing low-light image behavior. Under such conditions, non-convex mappings are required to stretch the dark regions while compressing the bright areas. This adjustment enhances shadow details and prevents overexposure in bright regions—critical for improving visual quality. To this end, we design a set of non-convex bases with \( f_0 = 1 \) and each \( f_k(x) \) is a polynomial of order \( k \) that passes through the points \( (0, 0) \) and \( (1, 1) \). These polynomials capture varying curvature patterns, from nearly linear to concave, effectively handling intensity redistribution in low-light scenarios, as shown in \cref{fig:nolinear}. The set of polynomial bases thus forms a low-dimensional manifold of valid tone-mapping operations \(\mathcal{F}(\cdot) \). 

Importantly, it is not an error-prone approximation for us to use \( \mathcal{F}(\cdot) \); instead, it constrains the nonlinear function to a finite \( n \)-dimensional polynomial, leveraging prior knowledge to balance expressiveness and computational efficiency. The transformed \( U = \mathcal{F}(I') \) provides the enhanced image for downstream tasks.

Our method shares similarities with Zero-DCE \cite{guo2020zero}, which also uses a neural network to generate coefficients for quadratic curve fitting. However, our approach differs by explicitly considering low-light behavior, preventing suboptimal convergence. Specifically, we use \( 3 \times 3 \) convolution layers to predict pixel-wise coefficient maps \( \{C_k\}_{k=0}^n \), which are then combined with the polynomial bases (as shown in \cref{fig:framework} (c)). To prevent gradient vanishing, skip connections are used, and \( x \) is subtracted from the original polynomial function to preserve the curve’s shape.

\subsection{Self-Boost Regularization}\label{method:sec3}

To enhance the compatibility between linear and nonlinear components in \cref{method:sec1}-\ref{method:sec2}, we propose a Self-Boost regularization mechanism. Given the linearly transformed image \( I' = P' \cdot I \) from the RAW Bayer input \( I \), the nonlinear module generates the enhanced image \( U = \mathcal{F}(I') \).

Ideally, given an oracle sRGB image \( \textcolor{camera}{U^*} \in \mathbb{R}^{3 \times H \times W} \) captured under normal lighting conditions, the optimal linear mapping can be found via the closed-form least squares solution:
\begin{equation}\label{eq9}
P^{*} = \mathop{\arg\min}\limits_{P'} \lVert \textcolor{camera}{U^*} - P' \cdot I \rVert ^2 = \textcolor{camera}{U^*} \cdot I^T \cdot (I \cdot I^T)^{-1}, 
\end{equation}
where \( P^{*} \in \mathbb{R}^{3 \times 4} \) and all images are reshaped into matrices of size $[C\times (H\times W)]$. 
However, this ideal formulation requires paired RAW-sRGB samples  $(I, \textcolor{camera}{U^*})$, the acquisition of which is prohibitive expensive. 

As a relaxation, we propose a self-supervised approximation by substituting the oracle \( \textcolor{camera}{U^*} \) with the nonlinear module's own output, \( U \). This is motivated by the intrinsic feature hierarchy hypothesis, which suggests that deeper network layers produce representations closer to the final objective than shallow layers \cite{bi2024learning}. Critically, because $U$ is a function of $P'$, the closed-form solution in \cref{eq9} is no longer holds when substituting $U$ for $\textcolor{camera}{U^*}$. Nevertheless, our objective is to encourage the linear component's output, $P' \cdot I$, to closely approximate the nonlinear output, $U$. We therefore treat $U$ as a pseudo-target and define an approximate linear mapping $\tilde{P}$ as follows:
\begin{equation}
    \tilde{P} := U \cdot I^T \cdot (I \cdot I^T)^{-1}.
    \label{eq:P_tilde}
\end{equation}

This formulation serves a dual purpose: it facilitates a more effective backpropagation of gradients from the high-level detection loss to the linear component $P'$, 
while also providing $U$ with greater optimization potential in the adversarial process of this regularization and detection loss.
We verify the validity of these properties in the appendix.


A direct alignment between \( P' \) and \( \tilde{P} \) is inappropriate because $\tilde{P}$ is merely an approximation of the solution $P^{*}$ and both matrices are continuously updated during training. Therefore, instead of enforcing a strict equality, we encourage directional consistency between the corresponding row vectors of \( P' \) and \( \tilde{P} \). We decompose each matrix into its constituent row vectors, \( P' = (\mathbf{p'_1}, \mathbf{p'_2}, \mathbf{p'_3})^T \) and \( \tilde{P} = (\mathbf{\tilde{p}_1}, \mathbf{\tilde{p}_2}, \mathbf{\tilde{p}_3})^T \), where each vector represents the linear projection from the four Bayer channels to a single RGB channel. Our Self-Boost loss, \( \mathcal{L}_{sb} \), is then defined as the sum of the cosine distances between these corresponding vector pairs:
\begin{equation}
\mathcal{L}_{sb} = \sum_{\mathbf{p'_i} \in P', \, \mathbf{\tilde{p}_i} \in \tilde{P}} \lVert \mathbf{1}-\cos(\mathbf{p'_i}, \mathbf{\tilde{p}_i})\rVert.
\end{equation} 

To prevent premature convergence, we activate $\mathcal{L}_{sb}$ after $N$ warmup epochs. The compound loss integrates with detection supervision as:
\begin{equation}
\mathcal{L} = \mathcal{L}_{det} + \lambda \cdot \mathcal{L}_{sb},
\end{equation}
where \( \lambda \) is empirically set to \( 1 \times 10^{-2} \) for all experiments. 

\section{Experiments}


\noindent \textbf{Implementation Details.} Our framework is based on the MMDetection toolbox \cite{chen2019mmdetection} and runs on a Tesla-V100 32GB GPU. For object detection, we use the RetinaNet detector \cite{lin2017focal} with a ResNet \cite{he2016deep} backbone initialized with ImageNet pre-trained weights. The model is trained with the SGD optimizer with an initial learning rate of 0.001, momentum of 0.9, and weight decay of 0.0001 for 15 epochs across all datasets. Random horizontal flipping is applied as augmentation, and all images are resized to $400 \times 600$. We choose the order $n=8$ for the nonlinear component in \cref{method:sec2}.

\noindent \textbf{Datasets and Metrics.} We evaluate our method on two real-world RAW datasets, LOD \cite{hong2021crafting} and NOD \cite{morawski2022genisp}, and a synthetic dataset, SynCOCO \cite{lin2014microsoft}. \textbf{The LOD dataset} contains 2,230 low-light RAW images from a Canon EOS 5D Mark IV, spanning 8 object classes. We use 1,800 for training and 430 for testing, evaluating performance with VOC-style mAP. \textbf{The NOD dataset} consists of RAW images captured in low-light environments with two cameras (Sony RX100 VII and Nikon D750), containing 3 object classes. The Sony dataset has 2,751 training and 321 test images, while the Nikon dataset has 3,206 training and 400 test images. In both datasets, each Bayer RAW image (.NEF, .ARW) is demosaiced into four RGBG channels, and resized to match the size of RGB images before being fed into the model. Performance is evaluated using COCO-style mAP, mAP\textsubscript{50}, and mAP\textsubscript{75}. 

Additionally, we generated a \textbf{synthetic COCO RAW dark dataset} using the widely used COCO dataset \cite{lin2014microsoft}. Following the pipeline from LIS \cite{chen2023instance}, we first apply the inverse ISP process to normal-light RGB images to generate synthetic RAW images. Dark noise from the physical noise model \cite{wei2020physics, Wei_Fu_Zheng_Yang_2021} is then injected to simulate dark RAW images. The same metrics as those used in COCO are applied for training and evaluation.



\begin{table}[t]
  \centering
  \caption{Detection performance comparison on the real-world LOD dataset. \textbf{Bold} indicates the best result, and \underline{underline} indicates the second best. $\ast$ denotes a two-stage training method.}
\small
    \begin{tabular}{cccc}
    \toprule
    \multirow{2}[2]{*}{Image Format} & \multirow{2}[2]{*}{Method} & ResNet18 & ResNet50 \\
          &       & mAP   & mAP \\
    \midrule
    \multirow{5}[2]{*}{\parbox[t]{5.5em}{\centering RGB \\ RAW-RGB}} & default ISP & 55.2 & 59.1  \\
          & demosaic & 52.6 & 61.3  \\
          & LIS \cite{chen2023instance}   & 50.5 & 60.8  \\
          & FeatEnHancer \cite{hashmi2023featenhancer} & 60.8 & 64.3  \\
          & RAW-Adapter \cite{cui2024raw} & 55.4 & 61.1  \\
    \midrule
    \multirow{7}[2]{*}{Bayer RAW} & default ISP & \underline{63.6}  & 67.3 \\
          & demosaic & 59.7  & 65.1 \\
          & LIS \cite{chen2023instance}  & 58.4  & \underline{67.9} \\
           & SID* \cite{chen2018learning}  & 61.2 & 64.7  \\
          & FeatEnHancer \cite{hashmi2023featenhancer} & 63.4  & 67.0  \\
          & RAW-Adapter \cite{cui2024raw} & 59.9  & 66.2 \\
          & \textbf{Our Dark-ISP}  & \textbf{64.9} & \textbf{70.4} \\
    \bottomrule
    \end{tabular}%
  \label{LOD}%
\end{table}%

\begin{table}[t]
  \centering
\caption{Detection performance comparison on the NOD dataset using RAW images captured in low-light conditions with two cameras: Sony RX100 VII 320 and Nikon D750. } 
\small
    \begin{tabular}{ccccc}
    \toprule
    Camera & Method & mAP   & mAP\textsubscript{50} & mAP\textsubscript{75} \\
    \midrule
    \multirow{7}[2]{*}{Sony} & default ISP & 28.3 & 51.0  & 29.2  \\
          & demosaic & 19.7 & 40.0   & 18.4 \\
          & LIS \cite{chen2023instance}  & 28.3 & 48.3 & 29.7 \\
          & SID* \cite{chen2018learning}   & 27.4 & 47.2 & 28.2 \\
          & FeatEnHancer \cite{hashmi2023featenhancer} & \underline{30.3} & \underline{52.1} & \underline{31.5} \\
          & RAW-Adapter \cite{cui2024raw} & 28.9 & 50.7 & 29.6 \\
          & \textbf{Dark-ISP(Ours)} & \textbf{31.5} & \textbf{53.4} & \textbf{32.2} \\
    \midrule
    \multirow{7}[2]{*}{Nikon} & default ISP & 27.4 & 47.5 & 28.5 \\
          & demosaic & 27.6 & 48.6 & 27.8 \\
        & LIS \cite{chen2023instance}  & 26.5 & 46.4 & 27.4 \\
        & SID*\cite{chen2018learning}   & 22.2 & 42.0  & 20.5 \\
          & FeatEnHancer \cite{hashmi2023featenhancer} & \underline{28.8} & \underline{48.9} & \textbf{30.8} \\
          & RAW-Adapter \cite{cui2024raw} & 28.3 & 48.2 & 28.7 \\
          & \textbf{Our Dark-ISP} & \textbf{29.9} & \textbf{50.9} & \underline{30.7} \\
    \bottomrule
    \end{tabular}%
  \label{NOD}%
\end{table}%

\begin{table}[t]
  \centering
    \caption{Detection performance comparison on the SynCOCO dataset, a synthetic RAW dark dataset with a larger sample size.}
    \small
    \begin{tabular}{ccccc}
    \toprule
    Input Format & Method    & mAP   & mAP\textsubscript{50} & mAP\textsubscript{75} \\
    \midrule
    \multirow{5}[2]{*}{\parbox[t]{5.5em}{\centering RGB \\ RAW-RGB}} & default ISP & 21.4  & 34.5 & 22.9 \\
          & demosaic & 19.6 & 31.7  & 20.5 \\
          & LIS \cite{chen2023instance}  & 16.5 & 17.7  & 17.5 \\
          & FeatEnHancer \cite{hashmi2023featenhancer} & 21.3 & 34.4  & 21.8 \\
          & RAW-Adapter \cite{cui2024raw} & 21.0  & 34.3 & 22.1 \\
    \midrule
    \multirow{7}[2]{*}{Bayer RAW} & default ISP & 21.5  & 35.9 & 23.1 \\
          & demosaic & 21.0  & 34.1 & 22.8 \\
          & LIS \cite{chen2023instance}   & 21.4  & 34.5 & 23.1 \\
          & SID* \cite{chen2018learning}   & 21.2  & 34.6 & 22.6 \\
          & FeatEnHancer \cite{hashmi2023featenhancer} & \underline{22.4}  & \underline{36.1} & \underline{23.9} \\
          & RAW-Adapter \cite{cui2024raw} & 21.7 & 34.9  & 23.1 \\
          & \textbf{Our Dark-ISP} & \textbf{23.1} & \textbf{37.7 } & \textbf{24.4} \\
    \bottomrule
    \end{tabular}%
  \label{COCO}%
\end{table}%


\subsection{Comparison with State-of-the-Art Methods}

We compare Dark-ISP with  the original Dark RGB and RAW images as well as several state-of-the-art methods: SID \cite{chen2018learning}, FeatEnHancer \cite{hashmi2023featenhancer}, LIS \cite{chen2023instance}, and RAW-Adapter \cite{cui2024raw}. 
Dark RGB images are obtained through the ISP pipeline in LED \cite{jin2023lighting} and are represented as "default ISP" in the experiment. Dark RAW images, which only undergo demosaicing and binning operations, are called "demosaic".
SID, a pioneering dark RAW image denoising approach, requires a pre-trained denoising network before object detection tasks. FeatEnHancer is an end-to-end module for enhancing dark RGB images in perceptual tasks. LIS and RAW-Adapter are end-to-end methods that process RAW-RGB images. All methods were evaluated using the same data augmentation and training configurations for a fair comparison.

\begin{figure*}[t]
\centering
\includegraphics[width=2\columnwidth]{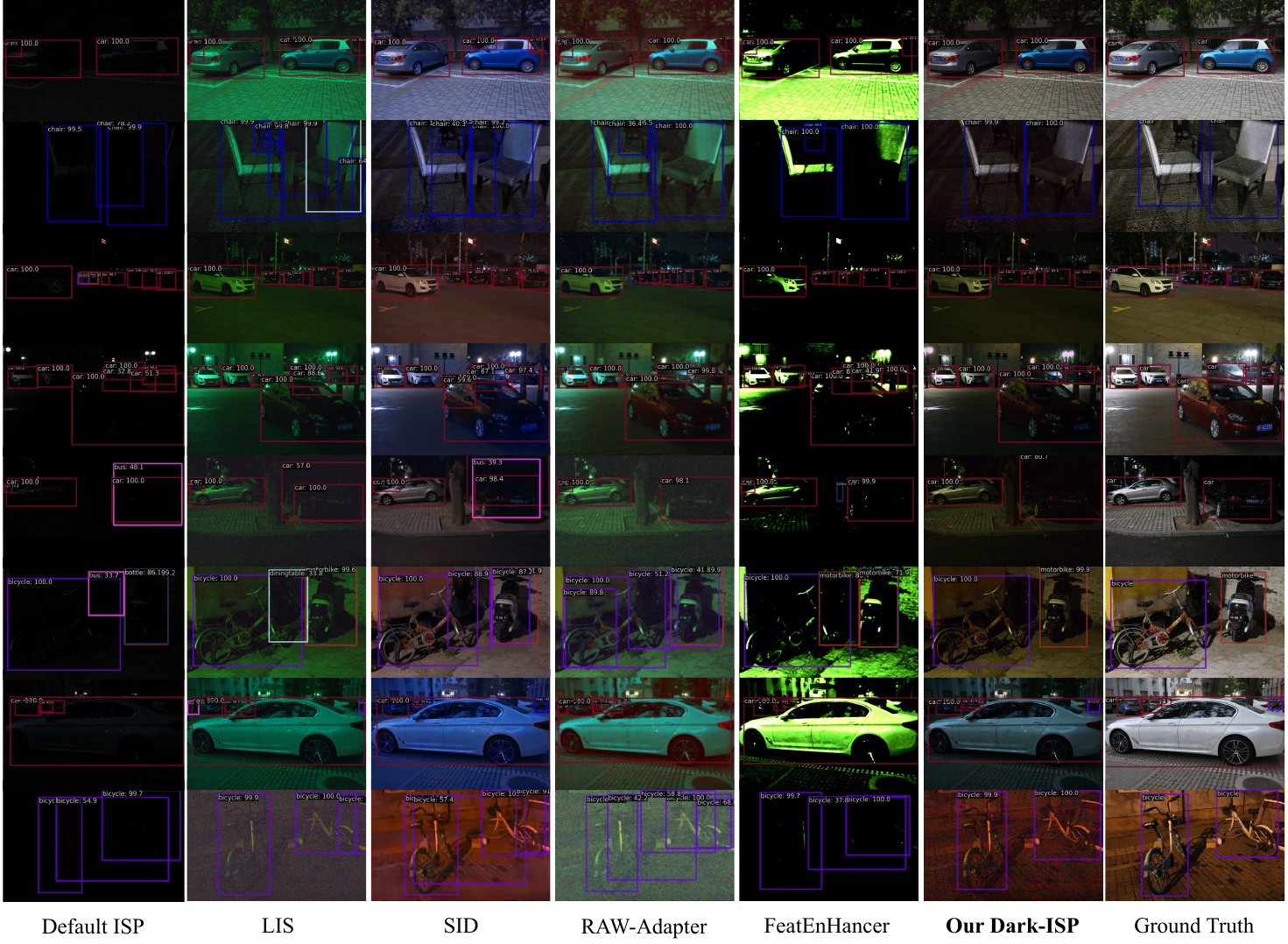} 
\caption{Visual comparisons on the LOD dataset. Results are shown on the enhanced images before being fed into the detection network backbone for each method. Our method outperforms others in minimizing missed and false detections, please zoom in for details.}
\label{fig:Vasual comparisons}
\end{figure*}

\subsubsection{Quantitative Analysis}

The detection performance on the LOD dataset is illustrated in \cref{LOD}. Our evaluation utilized two backbone depths (ResNet-18 and ResNet-50) to assess scalability. Inputs included both \textbf{Bayer RAW}, \textbf{RAW-RGB} and \textbf{RGB} images (low-bit-depth compressed versions of Bayer images), enabling a comparative study of image format impacts. As seen in the results, input from the Bayer RAW outperforms RGB or RAW-RGB images across most methods, highlighting that direct sensor data provides richer information.
Compared to other methods, our approach achieved the best results on both sizes of backbones (\textbf{64.9 mAP} and \textbf{70.4 mAP}), highlighting its ability to better leverage and extract useful information from Bayer RAW inputs in dark environment. 
It is important to note that SID, as a two-stage method, effectively reduces dark noise but shows minimal improvement in performance. This suggests that, for downstream tasks, the ability to adapt the RAW image distribution to the pre-trained backbone is crucial for achieving optimal performance. 
Furthermore, when using Bayer RAW as input, the basic camera default ISP performs on par with other learnable methods, indicating significant untapped potential in Bayer RAW inputs.



To evaluate the generalization of our method, we tested it on the NOD dataset (see \cref{NOD}). Our approach outperformed others on data from both Sony and Nikon cameras, demonstrating its robustness across diverse camera parameters and its ability to adapt to each camera’s characteristics using task-specific losses.


Finally, the results on the SynCOCO dataset, presented in \cref{COCO}, further validate the effectiveness of our method even when trained with a larger dataset. Despite significant differences between synthetic RAW images obtained through inverse ISP processing and real RAW images, the addition of extra channels provides the model with more data to learn from, and our method continues to effectively capture this information.

\subsubsection{Qualitative Analysis}

Qualitative results on the LOD dataset, as shown in Fig. \ref{fig:Vasual comparisons}, demonstrate that the proposed method consistently recalls most of the targets, even in challenging scenarios. Our approach excels at accurately detecting objects in dark environments compared to other methods. In terms of visual quality, the images enhanced by our proposed ISP are closer to the ground-truth normal RGB images, indicating that it has successfully learned the necessary transformations to align RAW domain images with the RGB domain of the pre-trained backbone.

\begin{figure}[t]
\centering
\includegraphics[width=1.02\columnwidth]{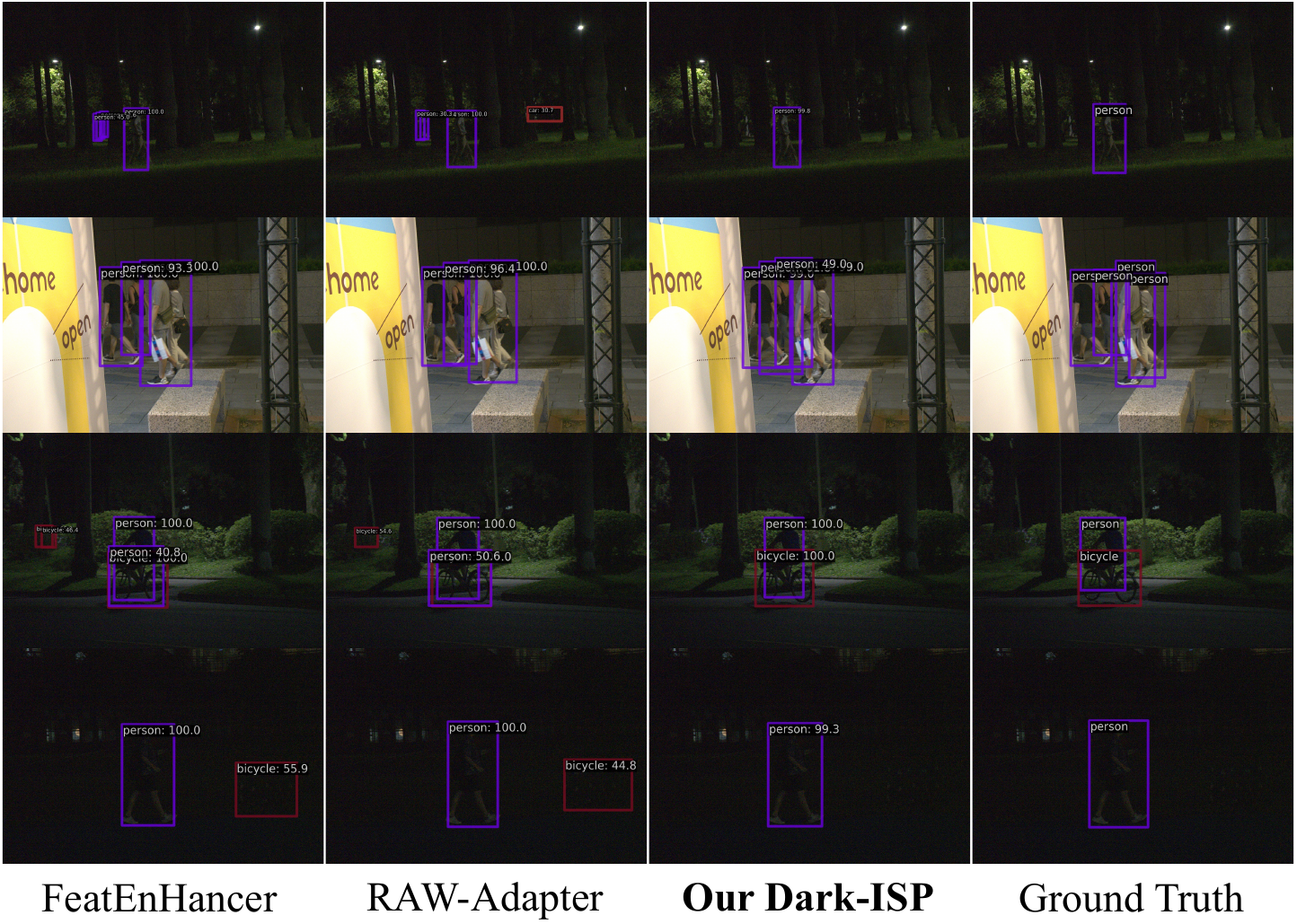} 
\caption{Visual comparison of our method with the two top-performing competitors on the NOD dataset (Sony camera).}
\small
\label{fig:NOD_Sony Vasual comparisons}
\end{figure}

\begin{figure}[t]
\centering
\includegraphics[width=1.02\columnwidth]{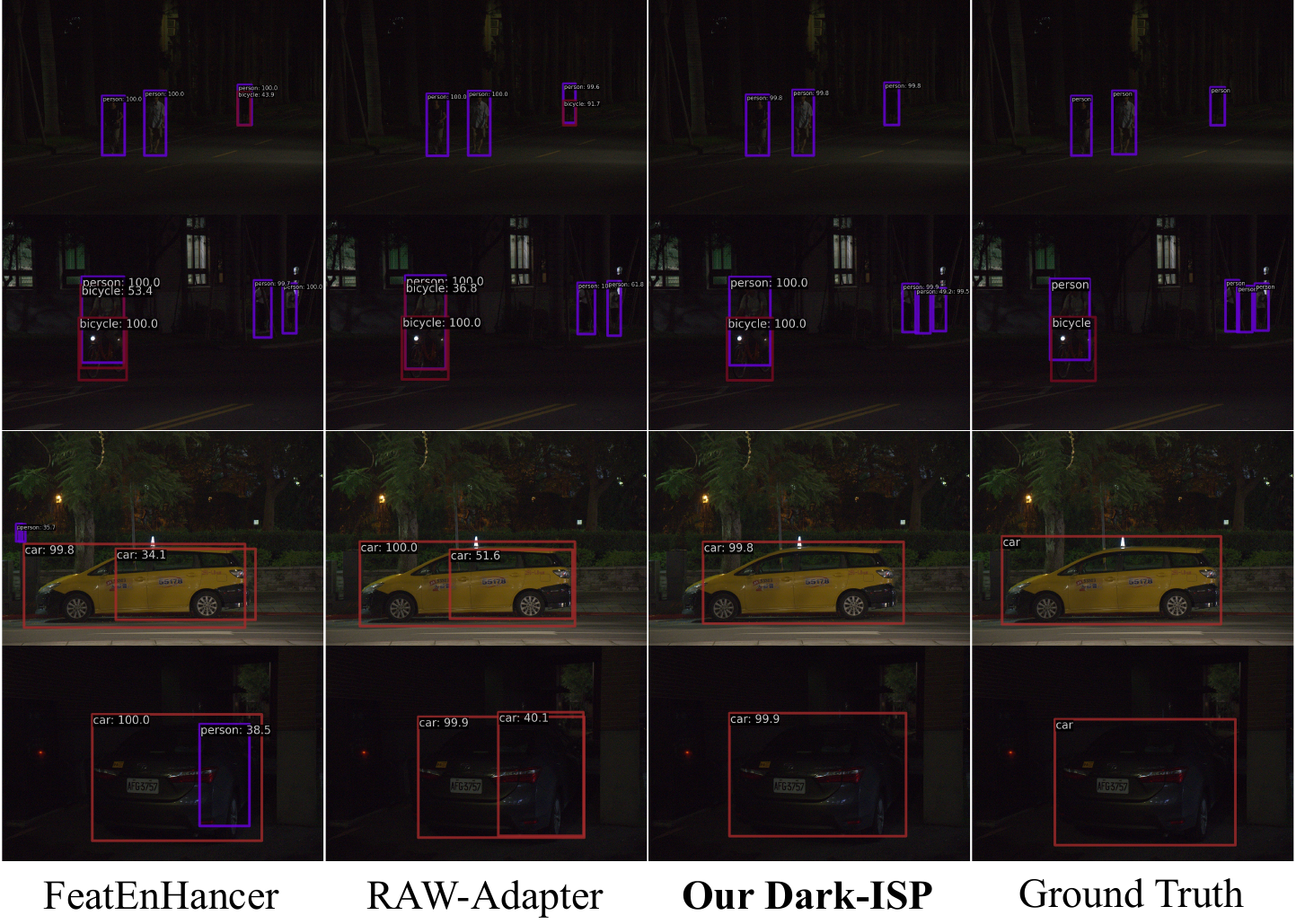} 
\caption{Visual comparison of our method with the two top-performing competitors on the NOD dataset (Nikon camera).}
\label{fig:NOD_Nikon Vasual comparisons}
\end{figure}

The results on the NOD dataset, visualized in Fig. \ref{fig:NOD_Sony Vasual comparisons} and Fig. \ref{fig:NOD_Nikon Vasual comparisons}, further emphasize the effectiveness of our method. Compared to two competitive approaches, our method reliably detects objects in dark regions across both camera datasets, avoiding false positives and missed detections. This underscores the superiority of our approach in detecting objects under low-light conditions and its robustness across different camera setups.

\subsection{Ablation Study}

We conduct a series of ablation experiments on the LOD dataset to investigate the impact of the modules in the proposed pipeline on object detection in dark environments.

As shown in \cref{Ablation pipline}, the three modules in the pipeline serve distinct roles. The nonlinear component, due to its non-convex nature, better adapts image pixel distributions for dark perception tasks, outperforming the linear component. When combined, the two components exceed their individual performances. The Self-Boost mechanism enhances module interaction, improving information extraction from RAW images and boosting detection accuracy to 70.4 mAP.

For the linear component, we compared using fixed parameters from the traditional ISP, as well as using only local or global information. By introducing only a small number of parameters (0.345MB), it outperforms the Default ISP by 4 mAP in low-light detection. All results and detailed analysis can be found in the supplementary material.



The nonlinear component plays a crucial role in tone mapping and adjusting image distribution for downstream detection tasks. There are several design alternatives for this component, which we evaluate in this study. First, we use gamma correction from traditional ISP as the baseline and compare it with the learnable gamma values (Gamma$\dagger$)  \cite{ljungbergh2023raw} and a 3D color lookup table (LUT) method \cite{zeng2020learning}. Additionally, we examine ResMLP, the color manipulation technique used in RAW-Adapter \cite{cui2024raw}.
As shown in \cref{Nonlinear Ablation}, our method outperforms the alternatives even without the skip connection, by leveraging a non-convex prior that is well-suited for low-light conditions. When the skip connection is added, it mitigates the vanishing gradient problem, allowing our method to achieve the best performance while maintaining a relatively small number of parameters.


\begin{table}[t]
  \centering
  \caption{Ablation study of the proposed pipeline.}
  \small
    \begin{tabular}{cccc}
    \toprule
    Linear & Nonlinear & Self-Boost  Regularization & mAP \\
    \midrule
    \checkmark     &       &       & 66.6  \\
          & \checkmark     &       & 67.1 \\
    \checkmark     & \checkmark     &       & 68.7  \\
    \checkmark     & \checkmark     & \checkmark     & \textbf{70.4} \\
    \bottomrule
    \end{tabular}%
  \label{Ablation pipline}%
\end{table}%


\begin{table}[t]
  \centering
  \caption{Ablation study for Nonlinear component. w/o Skip means the removal of the skip connection in the Nonlinear Component.}
  \small
  \setlength{\tabcolsep}{15pt} 
    \begin{tabular}{ccc}
    \toprule
    Method & mAP   & Param(MB) \\
    \midrule
     Gamma & 66.4 & - \\
    Gamma$\dagger$ \cite{ljungbergh2023raw} & 68.0  & - \\
    LUT  \cite{zeng2020learning}  & 67.8 & 0.192 \\
    ResMLP  \cite{cui2024raw}  & 67.0  & 0.049 \\
    Zero-DCE  \cite{guo2020zero} & 68.0  & 0.304 \\
    Ours w/o Skip & \underline{68.6} & 0.136 \\
    \textbf{Our Dark-ISP} & \textbf{70.4} & 0.136 \\
    \bottomrule
    \end{tabular}%
  \label{Nonlinear Ablation}%
\end{table}%

\section{Conclusion}

We present Dark-ISP, a novel approach to RAW image processing for low-light object detection, leveraging physics-guided modular decomposition. Our key innovation lies in deconstructing conventional ISP pipelines into trainable linear sensor calibration and nonlinear tone mapping modules, preserving RAW data integrity while introducing task-driven content awareness. Our extensive experiments across various real-world and synthetic low-light datasets demonstrate that Dark-ISP outperforms state-of-the-art methods with fewer parameters, offering a lightweight yet powerful solution. By bridging the gap between physics-informed image processing and machine perception, Dark-ISP paves the way for more robust and generalizable visual systems that excel in low-light conditions and beyond. This approach also opens the door to broader applications in other perceptual tasks, such as segmentation, tracking, and even end-to-end autonomous driving.

{
    \small
    \bibliographystyle{ieeenat_fullname}
    \bibliography{main}
}

\end{document}


\maketitle

\section{Glossary and Terminology}

\subsection{Image Signal Processing (ISP) Terms}

The role of the ISP is to render the photon data collected by the sensor into the desired image as perfectly as possible. In this section, we outline the steps involved in processing the images recorded by the sensor in a digital camera to produce the final image: 

\begin{enumerate}[label={(\roman*)}]
    \item \textbf{Camera Sensor}: The camera sensor consists of a two-dimensional grid of photodiodes, where each photodiode is a semiconductor device that converts photons (light radiation) into charge, corresponding to a single pixel in the image. Color filters are placed over these photodiodes to produce color. This arrangement of color filter arrays (CFA) is typically named after Bryce Bayer \cite{delbracio2021mobile}.
    \item \textbf{Pre-processing}: Initial operations applied to Bayer sensor data to correct inherent sensor limitations and prepare the image for further processing. This includes adjustments such as black level correction, which sets the baseline pixel value to zero, and lens shading correction, which compensates for vignetting effects caused by lens imperfections.
    \item \textbf{Noise Reduction}: Techniques employed to minimize unwanted random variations in pixel intensity, known as noise, which can degrade image quality. Noise reduction is crucial for enhancing visual quality and is closely related to exposure time and camera ISO settings.
    \item \textbf{Demosaicing}: The process of reconstructing a full-color image from incomplete color samples output by an image sensor overlaid with the Bayer pattern. This involves interpolating the missing color information to produce a complete RGB image.
    \item \textbf{White Balance}: A method used to adjust the colors in an image to match the perceived color of the scene, ensuring that objects that appear white in person are rendered white in the photo. This process compensates for the color temperature of the illumination source.
    \item \textbf{Color Space Transformation}: The conversion of image data from one color space to another. In digital imaging, this often involves mapping white-balanced pixel data to an intermediate color space (e.g., CIEXYZ) and then to a display-referred color space (e.g., sRGB), typically using 3×3 transformation matrices specific to the camera.
    \item \textbf{Color and Tone Correction}: Adjustments made to the color balance and tonal range of an image to achieve the desired visual appearance. These corrections are often implemented using 3D and 1D lookup tables (LUTs) and may include tone mapping to compress the dynamic range of the image.
    \item \textbf{Sharpening}: Techniques applied to enhance the perceived sharpness of an image by emphasizing edges and fine details. Methods such as unsharp masking or deconvolution are commonly used to achieve this effect.
\end{enumerate}

\subsection{RAW Image Formats}

\begin{itemize}
    \item \textbf{Bayer RAW image}: A Bayer image is a format that captures color information from the sensor arrangement of color filter arrays.
    \item \textbf{RGB-RAW image}: A demosaiced Bayer image is converted to a standard color space and stored in 8-bit RGB format(.png, .JPG, .JEPG).
    \item \textbf{sRGB image}: A standard RGB (Red, Green, Blue) color space created cooperatively by HP and Microsoft for use on monitors, printers, and the internet. It defines a specific range of colors that can be displayed, ensuring consistency across different devices.
\end{itemize}

\section{Supplementary Experimental Results}
This section provides additional experimental details and results not presented in the main paper, including performance evaluations and ablation studies.

\subsection{Detection Performance on LOD Dataset}

\cref{LOD_COCO} presents the detection performance results on the LOD dataset using COCO metrics. As shown, our method outperforms competing approaches in both mAP, mAP\textsubscript{50}, and mAP\textsubscript{75}, across both ResNet50 and ResNet18 backbones.

\begin{table*}[t]
  \centering
  \caption{Detection performance comparison on the real-world
LOD dataset for COCO metric. }
 \small
    \begin{tabular}{cccccccc}
    \toprule
    \multirow{2}[4]{*}{Image Format} & \multirow{2}[4]{*}{Method} &       & ResNet50 &       &       & ResNet18 &  \\
\cmidrule{3-8}          &       & mAP   & mAP 50 & mAP 75 & mAP   & mAP 50 & mAP 75 \\
    \midrule
    \multirow{5}[2]{*}{\parbox[t]{5.5em}{\centering RGB \\ RAW-RGB}} & default ISP & 52.2 & 77.9  & 58.1 & 45.9 & 73.9  & 49.5 \\
          & demosaic & 53.4 & 79.5  & 59.7 & 48.3 & 76.9  & 53.2 \\
          & LIS \cite{chen2023instance}   & 52.0  & 78.4  & 57.5 & 46.4 & 74.0  & 49.5 \\
          & FeatEnHancer \cite{hashmi2023featenhancer} & 53.9 & 80.5  & 58.5 & 50.0   & 78.2  & 55.3 \\
          & RAW-Adapter \cite{cui2024raw} & 53.3 & 79.8  & 58.8 & 47.5 & 75.7  & 52.5 \\
    \midrule
    \multirow{7}[2]{*}{Bayer} & default ISP & 58.4  & 83.7 & 65.9 & 53.2  & 80.5 & 58.9 \\
          & demosaic & 57.2  & 82.9 & 64.4 & 51.5  & 79.2 & 57.0 \\
          & SID \cite{chen2018learning}   & 56.5 & 82.5  & 63.0  & 51.9 & 79.7  & 57.5 \\
          & LIS \cite{chen2023instance}   & 57.8  & 83.5 & 65.6 & 52.4  & 79.8 & 58.2 \\
          & FeatEnHancer \cite{hashmi2023featenhancer} & 58.6  & 84.3  & 65.7 & 53.4  & 81.1  & 59.9 \\
          & RAW-Adapter \cite{cui2024raw} & 57.1  & 83.2 & 64.3 & 52.5  & 80.4 & 58.3 \\
          & Our Dark-ISP  & \textbf{58.9} & \textbf{84.4 } & \textbf{66.9} & \textbf{53.7} & \textbf{81.5 } & \textbf{60.9} \\
    \bottomrule
    \end{tabular}%
  \label{LOD_COCO}%
\end{table*}%

\begin{table}[t]
  \centering
  \caption{Ablation study for Linear component. Camera means the conventional White Balance and Color Space Transform operations in the ISP.
   Local indicates that we only allow the parameter matrix to perform Local Attention operations with the image. Global signifies that we only allow the parameter matrix to engage in Global Attention operations with the image. CCM refers to our approach of using only the $3\times3$ camera color correction matrix as the prediction target.}
   \small
    \begin{tabular}{ccc}
    \toprule
    Method & mAP   & Param(MB) \\
    \midrule
    Camera & 66.4 & -  \\
    Local & 67.0  & 0.168 \\
    Global & 68.6 & 0.177  \\
    Our Dark-ISP(CCM $3\times3$) & 69.2 & 0.343  \\
    \textbf{Our Dark-ISP(CCM $3\times4$)} & \textbf{70.4} & 0.345 \\
    \bottomrule
    \end{tabular}%
  \label{Linear Ablation}%
\end{table}%

\subsection{Ablation Study: Linear Component}

The Linear Component aims to learn a linear transformation that incorporates both global and local information from the image. We examine the impact of the two types of information on detection results, as shown in \cref{Linear Ablation}. The White Balance and Color Space Transformation operations from traditional ISP are used as baseline for comparison. Both types of information contribute to the learning of image color transformations. Compared to the local information (67.0 mAP), the global information (68.6 mAP) shows a more significant improvement. This may be because the linear operations in traditional ISP are inherently global operations, making the global information more compatible. Nevertheless, the information gains from the two are not coupled; thus, integrating them together leads to an additional performance improvement. Additionally, we also try using only the CCM matrix as the prediction target. Its size is 3x3, requiring the prediction of only 9 parameters, which makes its flexibility lower compared to our 3x4 joint matrix (12 parameters), resulting in a performance of 69.2 mAP.

\subsection{Ablation Study: Self-Boost  Regularization Starting Epoch}

\begin{figure}[h]
\centering
\includegraphics[width=0.45\columnwidth]{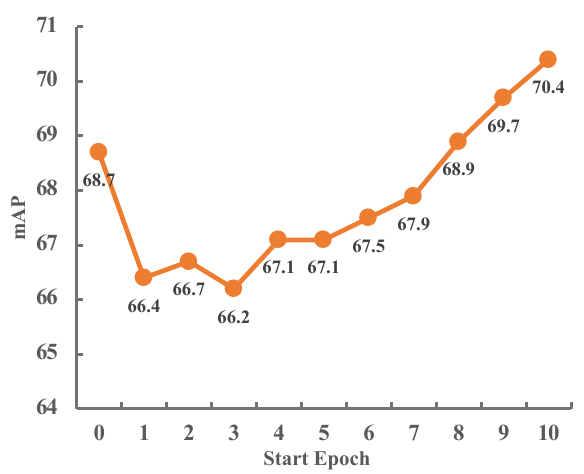} 
\caption{Ablation Study on the Starting Epoch of Self-Boost  Regularization.}
\label{fig:Loss Ablation}
\end{figure}

 We explore its impact on model convergence and performance. Self-Boost  Regularization embodies the idea that 'there is a hierarchy in learning and specialization in skills.' It allows advanced deep nonlinear features U to enhance the learning of weakly gradient-propagating shallow linear features $I'$, provided that $U$ possesses sufficient knowledge. In practice, Self-Boost  Regularization is activated after a few epochs of training to prevent $U$ from misleading $I'$ before it has adequately learned. Therefore, we conducted an ablation study on the activation epoch regarding this point, and the results are shown in \cref{fig:Loss Ablation}. It can be observed that at the beginning of training, due to $U$ not being fully converged and the randomness of the weights being high, the learning of $I'$ was misled, resulting in a performance drop compared to the baseline without this loss (\textbf{68.7 mAP}). As the activation epoch is pushed further back, $U$ becomes increasingly stable and starts to correctly guide $I'$, leading to a gradual performance increase. After activation at the 10th epoch, it reaches its peak at \textbf{70.4 mAP} .

\section{Nonlinear Function Design and Analysis}

\subsection{Impact of Tone Mapping on Image Representation}
\begin{figure}[t]
\centering
\includegraphics[width=\columnwidth]{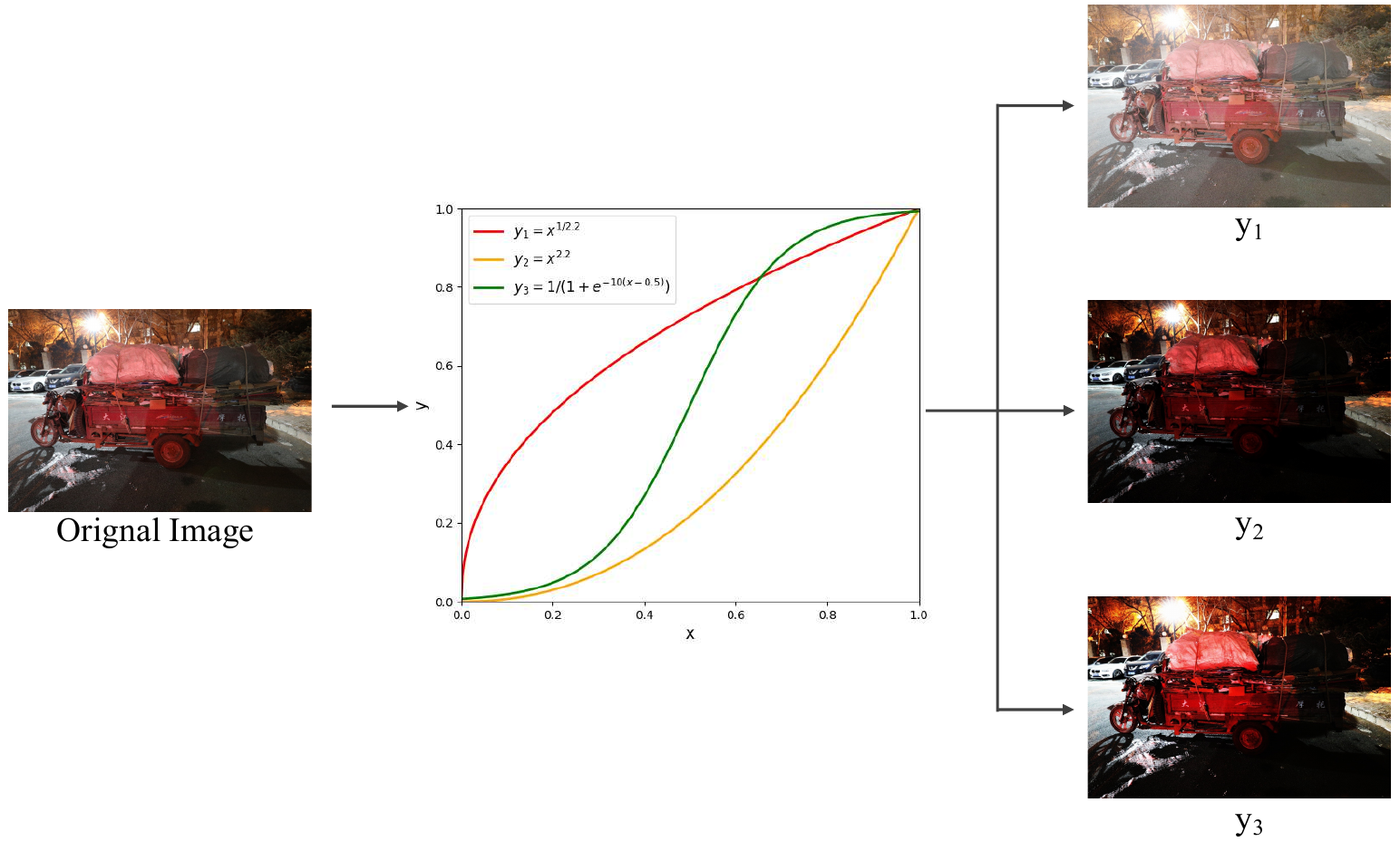} 
\caption{The impact of tone mapping functions with different shapes on images.}
\label{fig:mappingshape}
\end{figure}

In our study, we demonstrated the impact of tone mapping functions with different shapes on images. As shown in \cref{fig:mappingshape}, the concave function stretches the pixel values in the low-brightness regions while compressing the high-brightness areas, resulting in a brighter overall image. In contrast, the convex function has the opposite effect, making the originally dark regions even darker. The S-curve function further accentuates the shadow effects of objects, we believe it can also be beneficial for object detection tasks.

\subsection{Design of Non-Convex Polynomial Basis Functions}

\begin{figure}[t]
\centering
\includegraphics[width=\columnwidth]{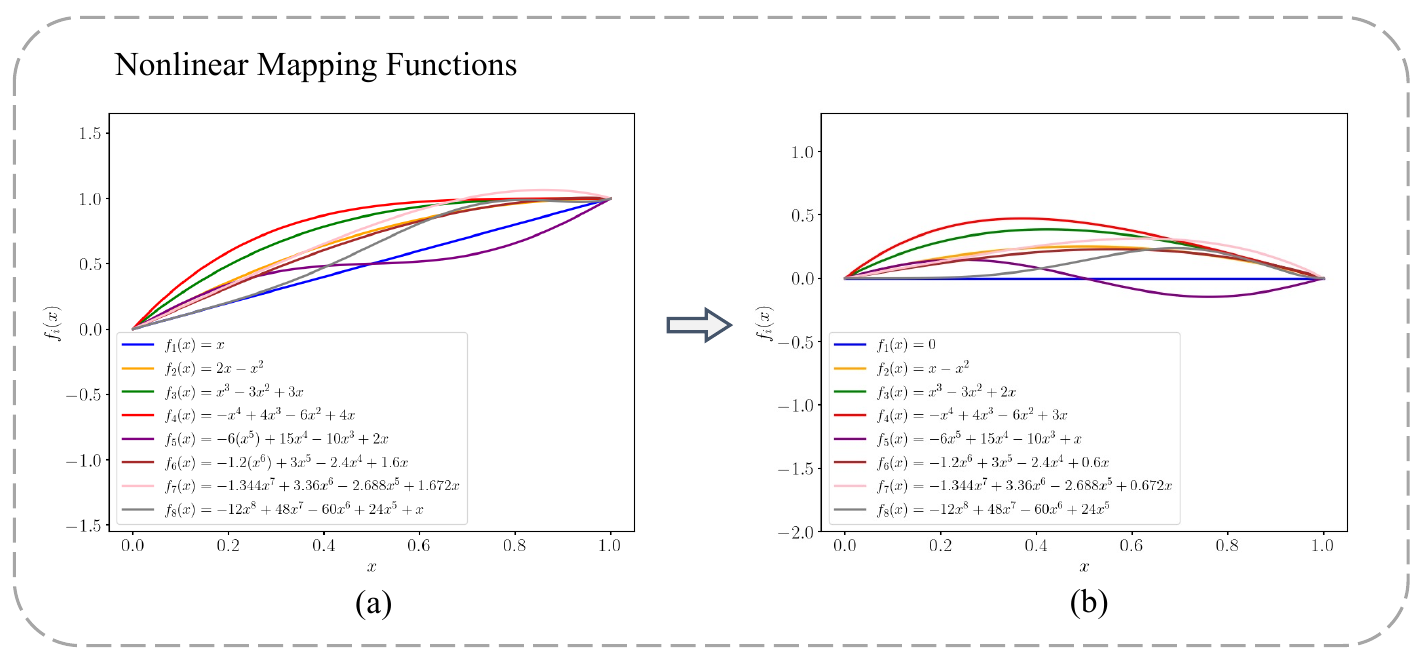} 
\caption{Specific functions of nonlinear mapping bases. (a) Predefined first to eighth order polynomial basis functions. (b) Subtracting $x$ to align with the skip connection and maintain curve shape.}
\label{fig:nonlinear mapping}
\end{figure}

Based on the above physical properties, we designed the basis functions in the nonlinear component. They satisfy the following properties: First, each function must pass through the points (0,0) and (1,1), representing the minimum and maximum brightness values of each pixel. Secondly, they must be non-convex functions on the interval [0,1] for dim image tones in low light environment. Therefore we selected polynomial functions of orders one to eight that satisfy the above two properties as bases to learn the ideal nonlinear mapping that best fits the detection task. The specific expressions of each basis function are shown in \cref{fig:nonlinear mapping}(a). We introduced skip connections to prevent the vanishing gradient problem caused by an increase in the number of network layers. To be compatible with this operation, each basis function needs to subtract $x$ from itself, and the resulting shape of each function are shown in \cref{fig:nonlinear mapping}(b).

\section{Discussion About Self-Boost Regularization}

\subsection{\texorpdfstring{The validity  of the $\tilde{P}$ definition}{The validity  of the P definition}}

\begin{figure}[h]
\centering
\includegraphics[width=0.8\columnwidth]{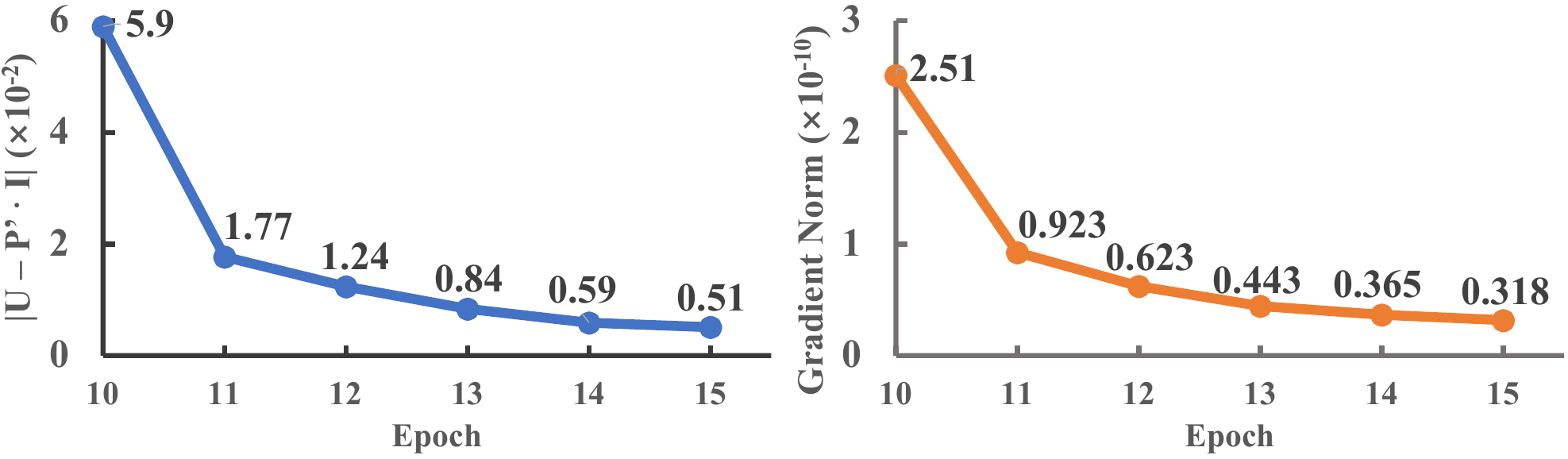} 
\caption{The trend of $|U - P'I| $ and the norm of the gradient.}
\label{fig:gradient}
\end{figure}

To verify the essence of the Self-Boost regularization derived from the defined equation 
\begin{equation}
    \tilde{P} := U \cdot I^T \cdot (I \cdot I^T)^{-1}
    \label{eq:P~},
\end{equation}
which ensures that $P' \cdot I$ approaches $U$ sufficiently, we calculated the average values of $|U - P' \cdot I|$ and the norm of $
\frac{\partial }{\partial P'}| U - P' \cdot I |^2
$ for each epoch after activating this mechanism from the 10th epoch during training. As shown in \cref{fig:gradient}, both values gradually converge to zero, demonstrating the effectiveness of this definition.
 
\subsection{Numerical Stability of the Self-Boost  Regularization}

In Section 3.3 of the main text, we use the inverse of \( (I \cdot I^T)^{-1} \) in Eq. (9) and Eq. (10) to compute the self-supervised signal for regularizing the learnable camera parameters \( P' \). However, when performing matrix inversion operations, it is essential to consider the condition number of the matrix \( I \cdot I^T \) to ensure numerical stability. This is because, when the condition number is large (e.g., approaching or exceeding \( 10^5 \) or higher), the matrix may become ill-conditioned, meaning it is numerically close to being singular or degenerate. In such cases, the inversion or pseudoinversion process can become unstable, leading to inaccurate results and potentially compromising our method.

From a theoretical perspective, \( I \cdot I^T \) is almost always invertible, as guaranteed by Sard's theorem. Consequently, \( I \cdot I^T \) is almost always positive definite, and its condition number is given by the ratio of the largest to the smallest eigenvalue:
\[
\kappa(A) = \frac{\lambda_{\max}}{\lambda_{\min}},
\]
where \( \lambda_{\max} \) and \( \lambda_{\min} \) denote the maximum and minimum eigenvalues of \( A \), respectively. For non-invertible cases, we resort to using \texttt{torch.linalg.pinv}, which internally computes the pseudoinverse via Singular Value Decomposition (SVD). This is a highly robust method, as SVD performs a numerically stable decomposition of the matrix. For matrices with large condition numbers, SVD reduces numerical errors by truncating small singular values, thus stabilizing the pseudoinverse computation.
This truncation effectively discards certain regularization directions, as it introduces numerical instability by neglecting smaller components. However, this does not significantly impact the final regularization of the camera parameters, as we are only selecting the more reliable directions for regularization.

\begin{figure}[t]
\centering
\includegraphics[width=0.45\columnwidth]{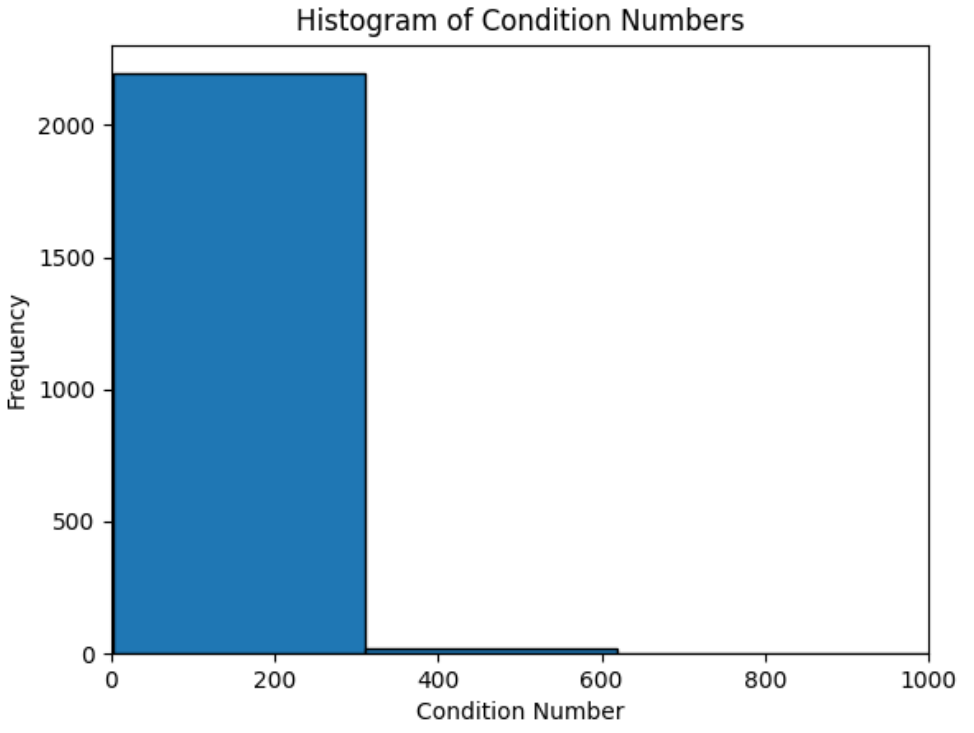} 
\caption{Histogram of the Condition Number of $I \cdot I^T $.}
\label{fig:condition}
\end{figure}

Additionally, we evaluated the condition number of \( I \cdot I^T \) for the images used in experiments. As shown in \cref{fig:condition}, the overall condition numbers of the matrices are not large, indicating that the inversion operations performed during our approach are numerically stable and do not pose significant risks for instability.

\section{Implementation Details}

\subsection{Architectural Details of Global-Local Attention}

\begin{figure}[t]
\centering
\includegraphics[width=1\columnwidth]{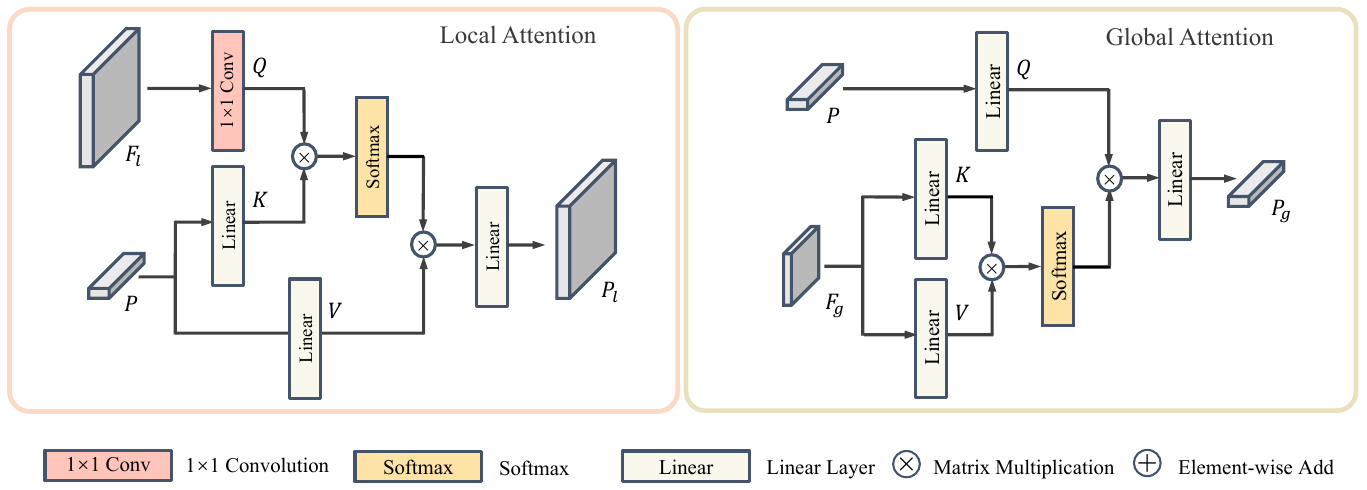} 
\caption{The Local Attention and Global Attention}
\label{fig:2}
\end{figure}

\cref{fig:2} provides a detailed visualization of the global-local attention mechanism used in our model. The figure illustrates how attention is applied locally and globally to enhance feature learning.

\subsection{Implementation of Comparative Methods}

This section details the implementation of baseline methods compared in the main paper, including FeatEnHancer and RAW-Adapter \cite{cui2024raw}. For a fair comparison, the implementation of all methods is consistent with the open-source libraries in the \textbf{github}. In the original method, LIS and RAW-Adapter \cite{cui2024raw} use RAW-RGB data, which represents the average of the two green channels of the Bayer image after compressing the bit depth and storing it in the sRGB color space. FeatEnHancer \cite{hashmi2023featenhancer} takes dark RGB as input directly, and SID \cite{chen2018learning} takes Bayer image as input directly. In our experiments, we also implemented the aforementioned comparative methods on data generated directly from Bayer images. For LIS \cite{chen2023instance} and RAW-Adapter \cite{cui2024raw}, we similarly averaged the two green channels as in LIS \cite{chen2023instance}, but without bit depth compression, directly passing the float-type tensor to the network. For FeatEnHancer \cite{hashmi2023featenhancer}, we processed the Bayer image using the same workflow as the default ISP used in the LED \cite{jin2023lighting} open-source library, obtaining a float-type representation of the RGB image before inputting it to the network.

\section{Efficiency  Analysis}

We conducted an efficiency analysis of each model on the LOD dataset, while also introducing two additional methods for comparison: ROAD\cite{xu2023toward} and IA-ISP\cite{liu2022image}. Both methods utilize the Bayer format as input for end-to-end training of the downstream task. As shown in \ref{tab:params}, Dark-ISP maintains high efficiency and low memory usage, with inference time comparable to other methods while achieving superior performance. Each module in our method effectively leverages rich prior knowledge, resulting in a lightweight ISP module that strikes an optimal balance between performance and efficiency, making it highly suitable for real-world applications.

\begin{table}[htbp]
  \centering
  \caption{Efficiency analysis for each method. The parameters exclusively count the modules apart from the downstream task detector.}
    \begin{tabular}{ccccc}
    \toprule
    Methods & mAP   & Params(MB) & Inference Time (ms) & GFLOPS \\
    \midrule
    SID\cite{chen2018learning}   & 64.7  & 29.60  & 3.48  & 97.91  \\
    LIS\cite{chen2023instance}   & 67.9  & 3.30  & 3.24  & 51.95  \\
    RAOD\cite{xu2023toward}  & 66.0  & 0.28  & 3.28  & 51.50  \\
    IA-ISP\cite{liu2022image}  & 67.0  & 0.63  & 3.56  & 51.75  \\
    RAW-Adapter\cite{cui2024raw} & 66.2  & 2.19  & 3.51  & 52.23  \\
    FeatEnHancer\cite{hashmi2023featenhancer} & 67.0  & 0.53  & 3.95  & 78.58  \\
    \textbf{Our Dark-ISP} & \textbf{70.4 } & 0.49  & 3.42  & 83.32  \\
    \bottomrule
    \end{tabular}%
  \label{tab:params}%
\end{table}%

\section{Additional Visual Comparisons}

\begin{figure}[t]
\centering
\includegraphics[width=\columnwidth]{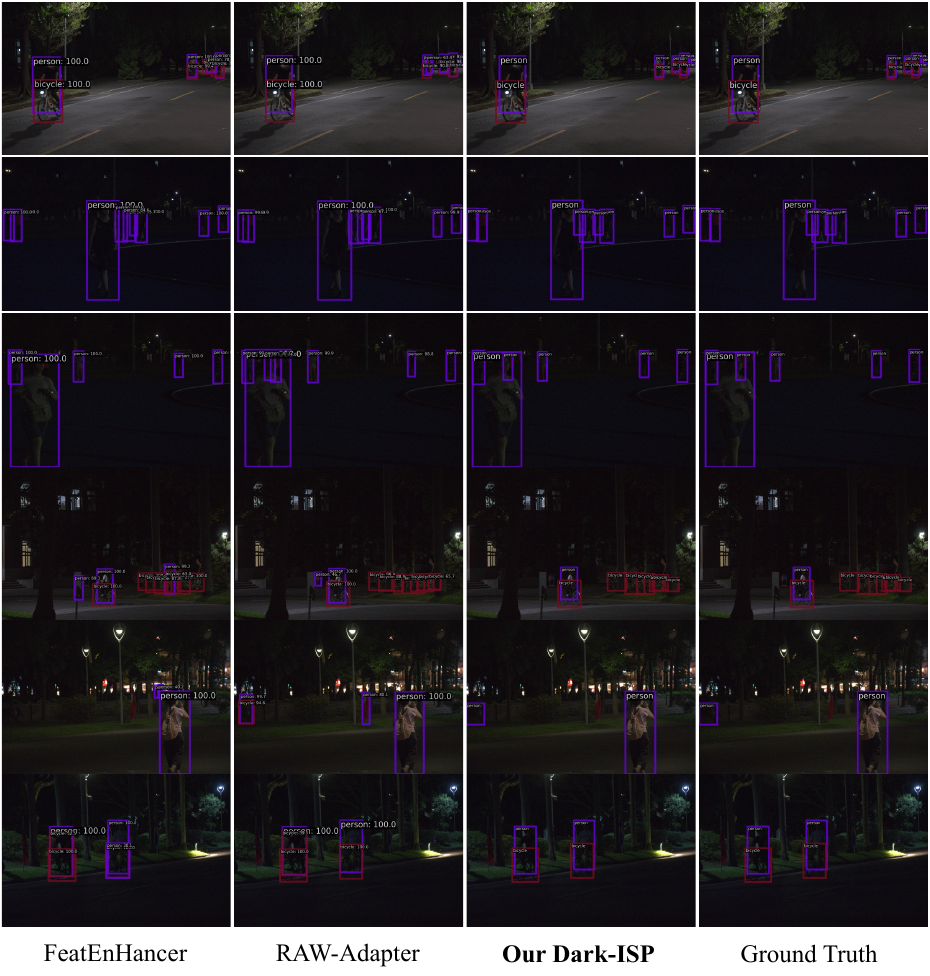} 
\caption{Visual comparisons on the NOD dataset.}
\label{fig:NOD1}
\end{figure}

\begin{figure}[t]
\centering
\includegraphics[width=\columnwidth]{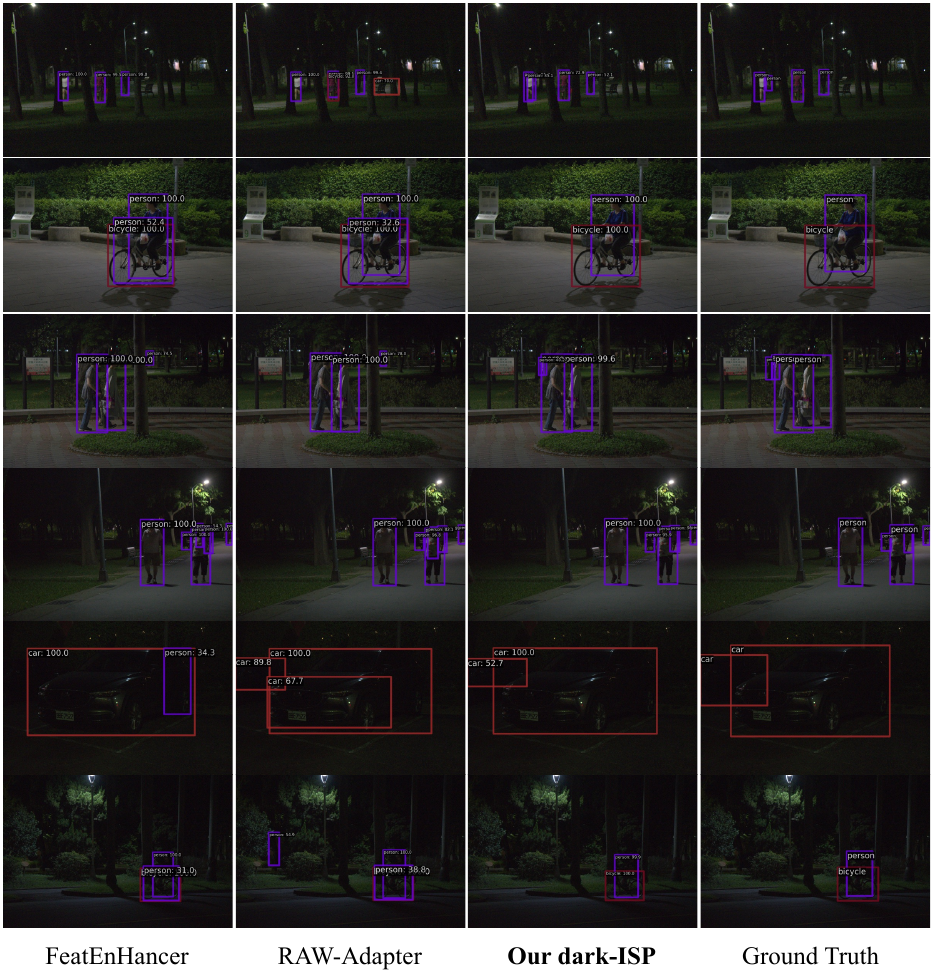} 
\caption{Visual comparisons on the NOD dataset.}
\label{fig:NOD2}
\end{figure}

This section includes additional qualitative comparisons of detection results on the NOD dataset. \cref{fig:NOD1} and \cref{fig:NOD2} demonstrate the qualitative performance comparison of Dark-ISP with other state-of-the-art methods (FeatEnHancer \cite{hashmi2023featenhancer}, RAW-Adapter \cite{cui2024raw}). Our method consistently detects more objects with higher accuracy in low-light conditions, avoiding false positives and missed detections.

\newpage

~

~

~

~

~

~

~

~

~

~

~

~

{
    \small
    \bibliographystyle{ieeenat_fullname}
    \bibliography{main}
}